\documentclass[draft]{agujournal2019}
\usepackage{url} 
\usepackage{lineno}
\usepackage{soul}

\usepackage{xcolor}

\draftfalse

\journalname{Journal of Advances in Modeling Earth Systems}

\begin{document}

%
%

\title{Controlled abstention neural networks for identifying skillful predictions for regression problems}

%
%
\authors{Elizabeth A. Barnes\affil{1} and Randal J. Barnes\affil{2}}
\affiliation{1}{Department of Atmospheric Science, Colorado State University, Fort Collins, CO, USA.}
\affiliation{2}{Civil, Environmental, and Geo- Engineering, University of Minnesota, Minneapolis, MN, USA.}




\correspondingauthor{Elizabeth A. Barnes}{eabarnes@rams.colostate.edu}



\begin{keypoints}
\item A simple neural network approach for adding uncertainty to climate regression problems is explored
\item A new abstention loss is introduced to identify, and preferentially learn from, more confident samples
\item The abstention loss outperforms other regression loss approaches for multiple climate use cases
\end{keypoints}

%
%

%
%

\begin{abstract}
The earth system is exceedingly complex and often chaotic in nature, making prediction incredibly challenging: we cannot expect to make perfect predictions all of the time. Instead, we look for specific states of the system that lead to more predictable behavior than others, often termed ``forecasts of opportunity''. When these opportunities are not present, scientists need prediction systems that are capable of saying ``I don't know.'' We introduce a novel loss function, termed ``abstention loss'', that allows neural networks to identify forecasts of opportunity for regression problems. The abstention loss works by incorporating uncertainty in the network's prediction to identify the more confident samples and abstain (say ``I don't know'') on the less confident samples. The abstention loss is designed to determine the optimal abstention fraction, or abstain on a user-defined fraction via a PID controller. Unlike many methods for attaching uncertainty to neural network predictions post-training, the abstention loss is applied during training to preferentially learn from the more confident samples. The abstention loss is built upon a standard computer science method. While the standard approach is itself a simple yet powerful tool for incorporating uncertainty in regression problems, we demonstrate that the abstention loss outperforms this more standard method for the synthetic climate use cases explored here. The implementation of proposed loss function is straightforward in most network architectures designed for regression, as it only requires modification of the output layer and loss function. 
\end{abstract}

\section*{Plain Language Summary}
The earth system is exceedingly complex and often chaotic in nature, making prediction incredibly challenging: we cannot expect to make perfect predictions all of the time. Instead, we can look for specific states of the system that lead to more predictable behavior than others, often termed ``forecasts of opportunity''. When these opportunities are not present, scientists need prediction systems that are capable of saying ``I don't know.'' We present a method for teaching neural networks, a type of machine learning tool, to say ``I don't know'' for regression problems. By doing so, the neural network focuses less on the predictions it identifies as problematic and focuses more on the predictions where its confidence is high. In the end, this leads to better predictions.

\clearpage

\section{Introduction}
The earth system is exceedingly complex and often chaotic in nature, making prediction incredibly challenging: we cannot expect to make perfect predictions all of the time. Instead, we look for specific states of the system that lead to more predictable behavior than others, often termed ``forecasts of opportunity''  \cite{Mariotti2020,Albers2019,Mayer2020,Barnes2020}. When skillful forecast opportunities are not present, scientists need prediction systems that are capable of saying ``I don't know.'' While this concept of forecasts of opportunity stems from weather and climate predictions, the general idea is far broader than this. For example, a forecast of opportunity framework may be beneficial when certain predictors are only helpful under certain circumstances. Additionally, if the predictor data has unknown errors or corrupted values (e.g. corrupted pixels in satellite imagery), a system that can say ``I don't know'' can act as an effective data cleaner: identifying the more skillful predictions, when they occur.

Many approaches to identify skillful forecasts of opportunity already exist. For example, retrospective analysis of the forecast can provide a sense of the physical circumstances that can lead to forecast successes or busts \cite<e.g.>{Rodwell2013-qz}, The ensemble spread can also give a sense of uncertainty in numerical weather prediction systems \cite<e.g.>{Van_Schaeybroeck2016-lo}. \citeA{Albers2019} used a linear inverse modeling approach to identify confident subseasonal predictions and showed that these more confident predictions were indeed more skillful. Recently, \citeA{Mayer2020} and \citeA{Barnes2020} suggested that machine learning, specifically neural networks, may be a useful tool to  identify forecasts of opportunity for subseasonal-to-seasonal climate predictions. Specifically, a classification network is first trained, then the predicted probabilities are ordered from largest to smallest. A selection of predictions with the highest probabilities are identified as possible forecasts of opportunity. While \citeA{Mayer2020} and \citeA{Barnes2020} show that this approach works well for classification tasks (i.e., predicting a specific category) where the network is already tasked with predicting a probability, it is less clear how one might apply this methodology to regression tasks (i.e., predicting a continuous quantity).

Most of the current machine learning approaches used to identify forecasts of opportunity, including those described above, are applied post-training. The network is first trained, and then the model confidence is assessed. Instead, here we build on the work by \citeA{Thulasidasan2019} and \citeA{Thulasidasan2020} to develop a deep learning abstention loss function for regression tasks that teaches the network to say ``I don't know'' (abstain) on certain samples \textit{during training}. 
The resulting controlled abstention network (CAN) preferentially learns from the samples in which it has more confidence and abstains on samples in which it has less confidence. The CAN is designed to identify the optimal abstention fraction, or abstain on a user-defined fraction via a PID controller; both approaches ultimately lead to more accurate predictions than our baseline approach. While alternative methods have recently been suggested for abstention (rejection) during training \cite{Geifman2019-paper,Geifman2019-thesis}, the CAN approach can be easily implemented in most any network architecture designed for regression, as it only requires modification of the output layer and loss function.

We demonstrate the behavior of the CAN on a simple 1D example, and then on synthetic climate data where the correct answer is known. We present two use cases with the climate data. The first use case explores the utility of the CAN to identify climate forecasts of opportunity and is modeled loosely after global teleconnections associated with the El Ni\~no Southern Oscillation \cite<e.g.>{McPhaden2006-pi,Yeh2018-tf}. The second use case explores the utility of the CAN to act as a data-cleaner by identifying input samples with corrupted pixels and preferentially learning on the uncorrupted samples.  

Section 2 introduces the synthetic climate data and general neural network architecture. Section 3 discusses the baseline loss function and the CAN in detail, and Section 4 presents the results. Additional discussion on the approach is provided in Section 5 and conclusions in Section 6.

\section{Data and Experiments}
\subsection{Synthetic climate data}
To demonstrate the utility of the controlled abstention network (CAN), we use the synthetic benchmark data set introduced by \citeA{Mamalakis2021}. While \citeA{Mamalakis2021} provides an extensive description of this data, we give a brief overview here. The data set consists of input fields $x_i$ and output series $y_i$ (where $i$ denotes the $i^{th}$ sample), which is a function of the input. The input fields represent monthly anomalous global sea surface temperatures (SSTs) generated from a multivariate normal distribution with a correlation matrix estimated from observed SST fields\footnote{https://psl.noaa.gov/data/gridded/data.cobe2.html}. The $i^{th}$ input sample consists of one map of SST anomalies, denoted as $x_i$. \citeA{Mamalakis2021} then defines the global response $y_i$ to sample $x_i$ as the sum of local, nonlinear responses. Specifically, 
\begin{linenomath*}
\begin{equation}
y_i = \sum_g F_g(x_i)
\end{equation}
\end{linenomath*}
where $g$ represents the grid point and $F_g$ is defined locally (at each grid point $g$) by a piecewise linear function. The slopes $\beta_n$ (where $n$ is an integer that runs from 1 to the number of piecewise linear segments, set here to 5) of each local function are chosen randomly from a multivariate normal distribution with correlation matrix, once again, estimated from observed SST fields. 
 
In the end, this data set consists of input maps of SSTs with spatial correlations indicative of observed SSTs, but where each input map is independent of the others. $y_i$ then represents the sum of contributions from each grid point across the globe, where that contribution is a nonlinear function (specifically, a piecewise linear function) of the SST value at that grid point. To speed up training time, we reduce the number of grid points (pixels) from that used by \citeA{Mamalakis2021} to 60 longitudes and 15 latitudes for a total of 900 grid points per input map. An example input map is shown in Fig. \ref{fig_arch}; its corresponding $y$ given in the title.

\subsection{Network architecture and training}
\begin{figure}
\begin{center}
\noindent\includegraphics[width=375px]{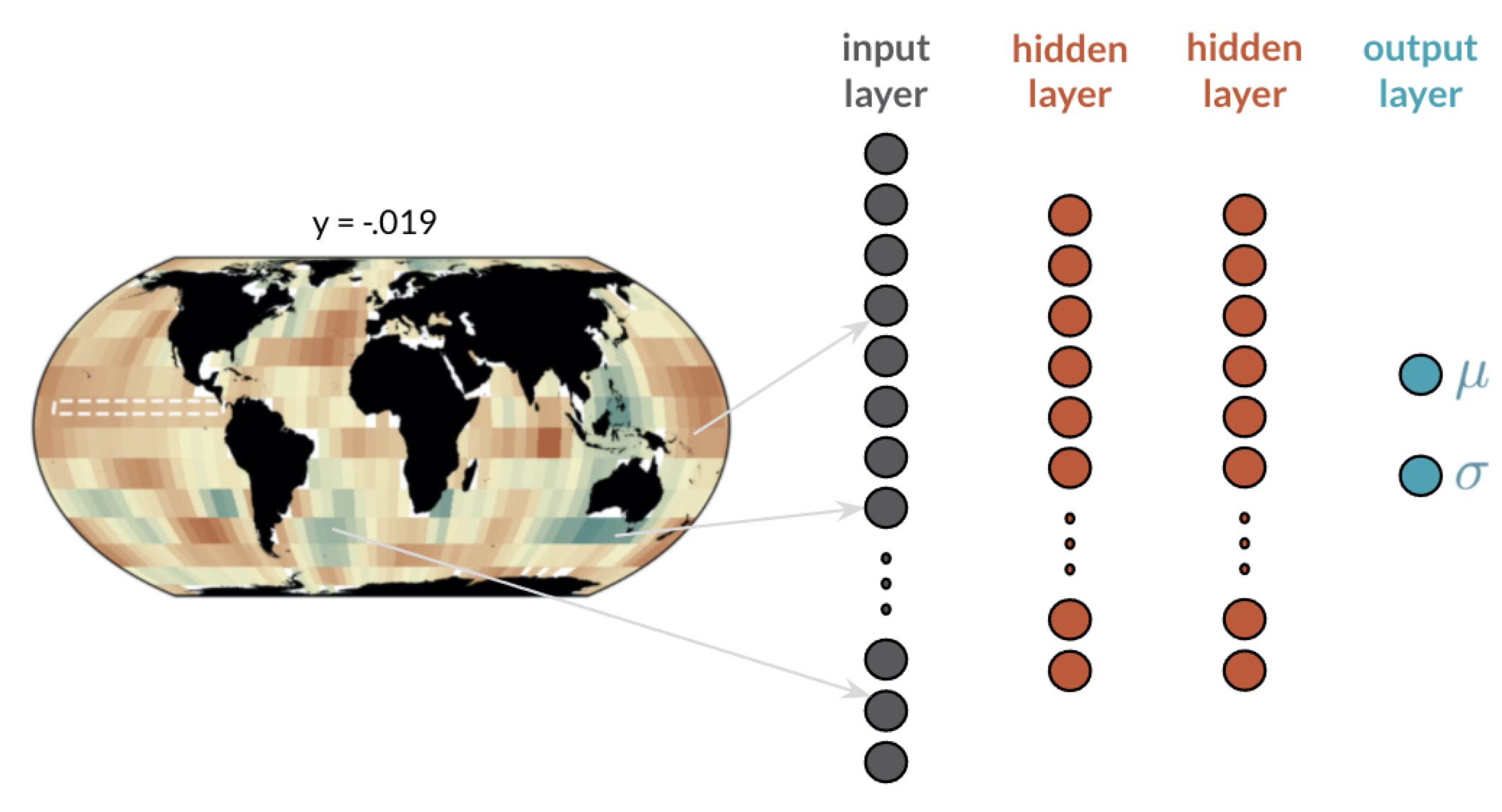}
\end{center}
\caption{General CAN architecture used for the experiments. A map of synthetic sea-surface temperature anomalies is fed into a fully connected network tasked with predicting $\mu$ and $\sigma$ for that sample.}
\label{fig_arch}
\end{figure}

For regression problems, it is typical to have a single output unit that provides the prediction by the network. Here, we add uncertainty estimates to our regression network by simply adding an additional output unit. We give these two output units the names $\mu$ and $\sigma$ as shown in Fig. \ref{fig_arch}. $\mu$ denotes the predicted value while $\sigma$ denotes the uncertainty related to that prediction. As we will show, we can take our interpretation even further and say that the network outputs a probabilistic prediction (conditional probability distribution) for the $j^{th}$ sample in the form of a normal distribution with mean $\mu_i$ and standard deviation $\sigma_i$.

We train a fully connected feed-forward network with two hidden layers with 50 and 25 units, respectively. As described above, the output layer consists of two units. We train with a ReLU (rectified linear unit) activation function, learning rate of 0.0005, and batch size of 32. Since the second output unit (denoted by $\sigma$ in Fig. \ref{fig_arch}) cannot be negative, we constrain it to being positive through the network setup. We train on 8,000 samples, validate on 5,000 samples, and test on 5,000 samples. While we could train on a much larger data set, we have intentionally kept the sample size relatively small to demonstrate the utility of the CAN when the sample size is relatively low --- as is the case for many geoscience applications. All quantities and figures are computed from the testing data unless otherwise specified.

We employ early stopping to automatically determine the optimal number of epochs to train. Specifically, the network stops training when the validation loss stops decreasing, with a {\tt patience} of 60 epochs. The network with the best performance on the validation loss is saved. Specifically for the CAN, we select the best performing network from epochs after the spin-up period, but we only consider epochs where the validation abstention fraction is within 0.1 of the abstention setpoint. For all examples shown here, 20 different networks are trained for each configuration (i.e., baseline ANN and CAN) by varying the randomly initialized weights.

The network was trained using Python 3.7.9 and TensorFlow 2.4.

\section{Methods}
\subsection{Baseline network with log-likelihood loss}
The baseline deep neural network (ANN) has the architecture shown in Fig. \ref{fig_arch} and trains using the negative log-likelihood loss defined for sample $x_i$ as
\begin{linenomath*}
\begin{equation}
\mathcal{L}(x_i) = -\log p_i. \label{loss_base}
\end{equation}
\end{linenomath*}
where $p_i$ is the value of the probability density function of a normal distribution ($\mathcal{N}$) with mean $\mu_i$ and standard deviation $\sigma_i$:
\begin{linenomath*}
\begin{equation}
p_i = \mathcal{N}(y_i,\mu_i,\sigma_i).
\end{equation}
\end{linenomath*}
This baseline model predicts $\mu_i$ and $\sigma_i$ for each sample, where $\mu_i$ is the model's best guess of $y_i$ and $\sigma_i$ is the associated uncertainty \cite<e.g.,>[Section 5.3.2]{Duerr2020}. 

Once the network is trained, we can invoke abstention on the less certain predictions by thresholding on $\sigma$ \cite<e.g.>{Mayer2020}. For example, the 20\% most confident predictions are determined as the smallest 20\% $\sigma$ values (or the $20^{th}$ percentile of the predicted $\sigma$). As we will show, this thresholding approach for abstention is itself very powerful and can be used as a simple way to add uncertainty to regression networks. In addition, this baseline approach will serve as a comparison for the CAN.
  
As an additional baseline, we will also compare our results with those obtained by training a standard feed-forward network containing a single output unit and a loss function defined by the mean absolute error (MAE). In this case, the network does not quantify uncertainty (i.e., $\sigma$). Consequently, only summary statistics over all testing predictions are provided.
  
Throughout this paper, we use ``coverage'' to denote the fraction of samples for which the network makes a prediction, and ``abstention'' to refer to the fraction of samples for which the network does not make a prediction. Thus, the percent coverage is always 100\% minus the percent abstention. For the baseline approach, abstention and coverage is computed post-training based on the predicted uncertainties $\sigma$ while for the CAN, these quantities are determined during the training itself (see next section).

\subsection{Controlled Abstention Network (CAN)}
\subsubsection{Abstention loss}
Unlike the baseline ANN, the CAN loss is designed to identify the less confident predictions so as to preferentially learn from the more confident predictions. The CAN loss for sample $x_i$ is defined as
\begin{linenomath*}
\begin{equation}
\mathcal{L}(x_i) = -q_i\log p_i - \alpha \log q_i . \label{loss}
\end{equation}
\end{linenomath*}
where $\alpha$ controls the amount of abstention (see next subsection) and $q_i$ represents the prediction weight defined as
\begin{linenomath*}
\begin{equation}
q_i =  \min\left (1.0, \left[\frac{\kappa}{\sigma_i} \right]^2 \right).
\end{equation}
\end{linenomath*}
$\kappa$ is a data-specific scale (see below). The prediction weight $q_i$ tells the CAN how much it should consider sample $j$ when it reduces the total loss during backpropagation. Note that Eq. \ref{loss} is very similar to the abstention loss of \citeA<>[Chapter 4]{Thulasidasan2020} and \citeA{BarnesBarnes2021Class} for classification networks.

The loss above works by increasing $\sigma_i$ values on samples that the CAN identifies as less certain. In this way, one can define abstention based on a threshold $\sigma$. Specifically, we define abstention by the CAN when the predicted $\sigma_i > \tau$. To define $\tau$, let $\mathcal{P}_m$ denote the $m^{th}$ percentile of the predicted validation $\sigma$ at the end of the spin-up period. Then $\tau = \mathcal{P}_m$ where $m$ is the percent coverage setpoint. For example, for a coverage setpoint of 80\% (abstention setpoint of 20\%), $\tau$ is set to the $80^{th}$ percentile of predicted validation $\sigma$ at the end of the spin-up period: $\tau = \mathcal{P}_{80\%}$. Note that since $\tau$ is defined by the validation data at the end of the spin-up period, it remains fixed during training and evaluation of the testing data.

We define $\kappa = \mathcal{P}_{90\%}$. This definition of $\kappa$ is something that the user can modify. For example, setting $\kappa=\tau$ is an obvious choice. However, we found that setting $\kappa = \mathcal{P}_{90\%}$ outperformed $\kappa=\tau$ and worked for all experimental setups here; consequently, we did not explore further tuning of this parameter. 

To summarize this section, the abstention loss looks a lot like the baseline loss (Eq. \ref{loss_base}). The main difference is the use of an additional scaling factor $q$ and an additional term that penalizes the network for large $\sigma$ predictions. This penalty is modulated by $\alpha$. $\kappa$ and $\tau$ are parameters set by the network during the spin-up period. $\kappa$ acts as a scaling parameter on $\sigma$ within the loss function. Samples with $\sigma$ larger than $\kappa$ contribute less to the loss function, while samples with $\sigma$ smaller than $\kappa$ contribute their full amount. $\tau$, on the other hand, sets the threshold used to define abstention and is used by the PID controller (see next section) when the user wishes to set a target coverage fraction.

\subsubsection{Setting the abstention setpoint}
The abstention loss, as defined in Eq. \ref{loss}, can be used in two distinct ways, depending on how $\alpha$ is determined. The first way is to set $\alpha$ to a predetermined constant. By doing this, the network is penalized equally throughout training for assigning high $\sigma$ values. If $\alpha$ is chosen correctly, the network can learn the optimal coverage percent from the data set. When $\alpha$ is held constant, the coverage setpoint is not set by the user and so we set $\tau = \kappa$. Physically, this represents the fact that the definition of abstention is set by the $90^{th}$ percentile of the predicted validation $\sigma$ values at the end of spin-up (i.e., $\mathcal{P}_{90\%}$). This works well because this same value is also used to define $\kappa$, the normalization factor used to set the confidence $q$ in Eq. \ref{loss}. 

Alternatively, $\alpha$ can be adaptively modified throughout training so that the network abstains on a specified fraction of the training samples. Inspired by the success reported in \citeA<>[Chapter 4]{Thulasidasan2020}, we implement a discrete-time PID controller (velocity algorithm) to modulate $\alpha$ throughout training \cite<e.g,>[Eq. (1.38)]{Visioli2006}.

\citeA{Thulasidasan2020} solely explores low abstention setpoints (e.g. 10\%), and evaluates the PID terms batch by batch. For our applications, however, we need the algorithm to work well for a broad range of abstention setpoints (e.g. from 10\% to 90\%). With a high abstension setpoint, say 90\%, and a batch size of 32, only 3 samples on average would be covered per batch --- this leads to to unstable behavior. Because of this, we evaluate the PID terms on 6 consecutive batches ($32 \times 6 = 192$ samples) which leads to more stable behavior of abstention fraction while not being so big as to impede training. Fig. \ref{fig_epochs} shows examples of the PID controller modulating $\alpha$ to control the abstention setpoint during training.

The training of the CAN occurs in two stages:
\begin{itemize}
    \item \textbf{Spin-up:} For the first $N_{spin}$ epochs, the CAN is trained using the baseline loss function given in Eq. \ref{loss_base}. At the end of spin-up, $\mathcal{P}_m$ is computed on the validation samples for $m$ between 10 and 90 in increments of 10.
    
    \item \textbf{Abstention training:} The CAN continues from where it stopped during the spin-up stage, but now trains using the abstention loss of Eq. \ref{loss}, with $\kappa$ and $\tau$ defined from $\mathcal{P}_m$. During this stage, $\alpha$ is either updated by the PID controller, or held constant at a user-defined value.
\end{itemize}
Based on these stages of training, there are only one to two \textit{new} free parameters to be determined by the user, depending on whether the PID controller is used to update $\alpha$ or whether $\alpha$ is held constant. Specifically, the user must choose the number of spin-up epochs, $N_{spin}$, for both methods and must also choose $\alpha$ if it is held fixed. While other parameters can certainly be tuned, we did not find it necessary for the range of experiments included in this paper.

\section{Results}
\subsection{A simple 1D example}
\begin{figure}
\begin{center}
\noindent\includegraphics[width=400px]{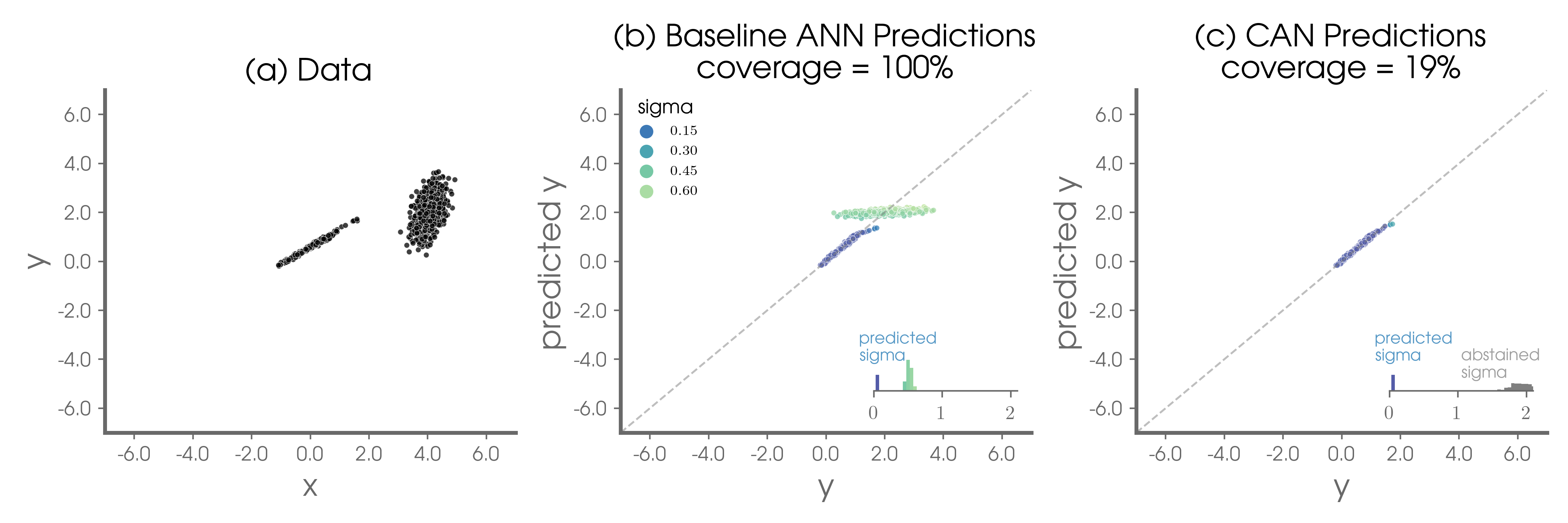}
\noindent\includegraphics[width=400px]{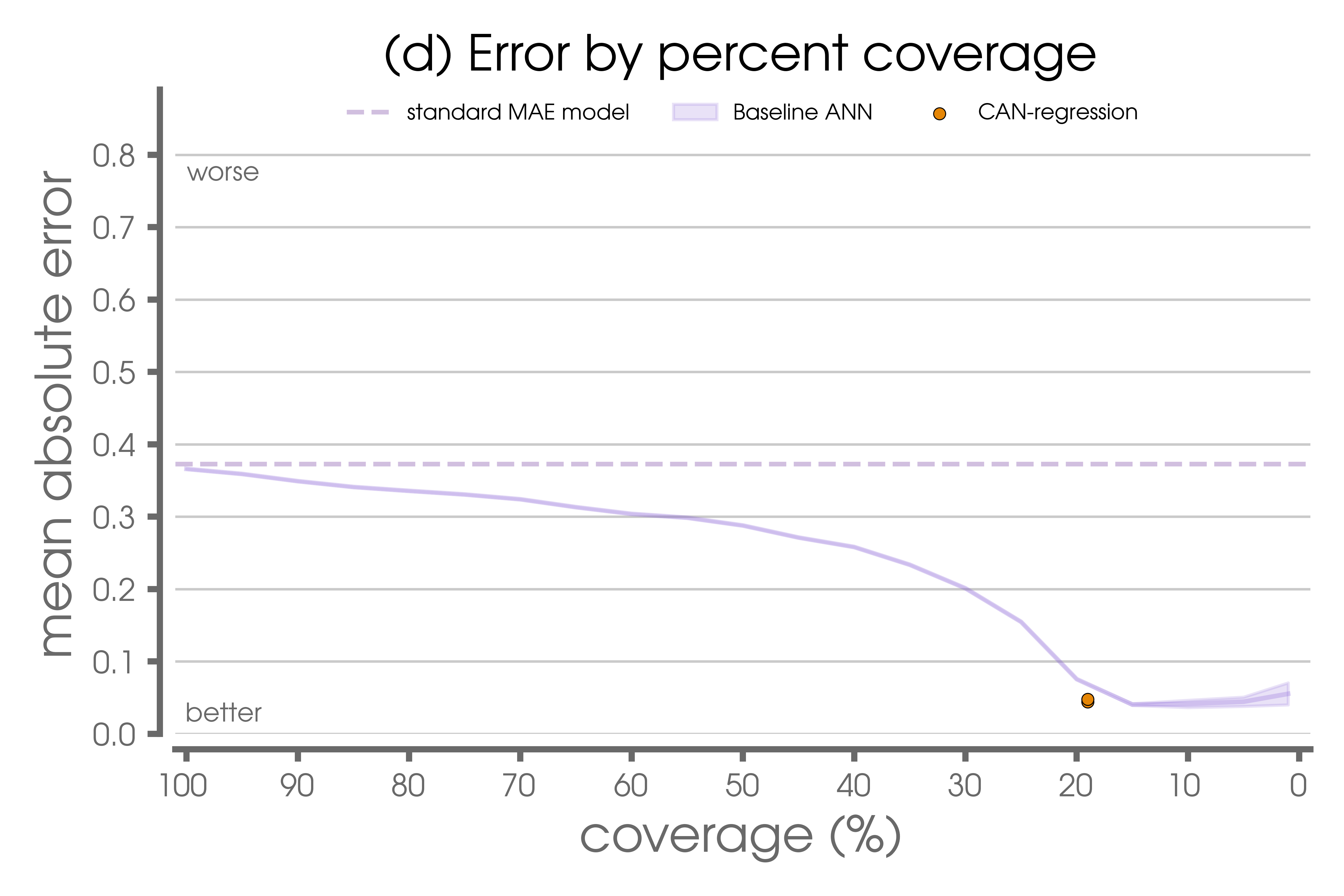}
\end{center}
\caption{\textbf{1D Example w/ Constant $\alpha$.} (a) Data used for the simple 1D example. (b) Predicted $y$ versus the true $y$ for the baseline ANN predictions. The dashed line denotes the one-to-one line --- a perfect prediction. (c) As in (b) but for the CAN predictions. Scatter plots only show covered predictions (i.e., non-abstained). Colors denote the predicted $\sigma$, and insets in (b,c) display histograms of the predicted $\sigma$ for both covered and abstained predictions. (d) Mean absolute error versus coverage for different neural network loss functions over a range of initialization seeds for constant $\alpha=0.1$. Purple shading denotes the full range of errors over 20 baseline ANN models;  the solid purple line denotes the median.}
\label{fig_ols_summary}
\end{figure}

Before we discuss results with the synthetic climate data, it is informative to explore the behavior of the baseline ANN and CAN for a simple example with a 1-dimensional input. Specifically, we define an $(x,y)$ data set as (Fig. \ref{fig_ols_summary}a):
\begin{linenomath*}
\begin{eqnarray}
x_c &=& \epsilon(4.0,.25)   \\
y_c &=& 1.0x_c - 2.0 + \epsilon(0.0,.5) \nonumber \\
x_l &=& \epsilon(0.0,.5)   \nonumber \\
y_l &=& 0.7x_l + 0.6 + \epsilon(0.0,.05) \nonumber \\
(x,y) &=& (\{x_c,x_l\},\{y_c,y_l\}) \nonumber
\end{eqnarray}
\end{linenomath*}
where $\epsilon(a,b)$ denotes a random variable drawn from a normal distribution with mean $a$ and standard deviation $b$. The data is created such that 30\% of the samples exist along the line (i.e. $(x_l,y_l)$), and 70\% of the points exist within the cloud (i.e., $(x_c,y_c)$). Fig. \ref{fig_ols_summary}a shows the data with $x$ on the x-axis and $y$ on the y-axis. The data is designed such that for $x$ less than about 2.5, the data largely follows a straight line with little noise. For larger $x$, the data shows a cloud of points with no clear linear relationship. Naively fitting a straight line through all of this data would result in a fit that performs poorly on most samples. Instead, we would like a network to predict the samples along the line with accuracy while also identifying the samples within the cloud as being highly unpredictable (``I don't know."). 

The network is trained to take the input value $x_i$ and predict $y_i$. For this 1D simple example only, we train a fully connected network with 2 hidden layers of 5 units each. We found that this architecture is complex enough to learn the linear fit but not so complex as to learn a separate fit for the cloud. The network is trained with a constant $\alpha$ to evaluate whether the CAN is able to identify the correct coverage fraction of 30\% and abstain on the remaining 70\%. We found that $\alpha=0.1$ works well. We set the number of spin-up epochs to $N_{spin} = 225$ and use a learning rate of 0.0001. Finally, we train on 3,000 samples, validate on 1,000 samples, and test on 1,000 samples.

\begin{figure}
\begin{center}
\noindent\includegraphics[width=200px]{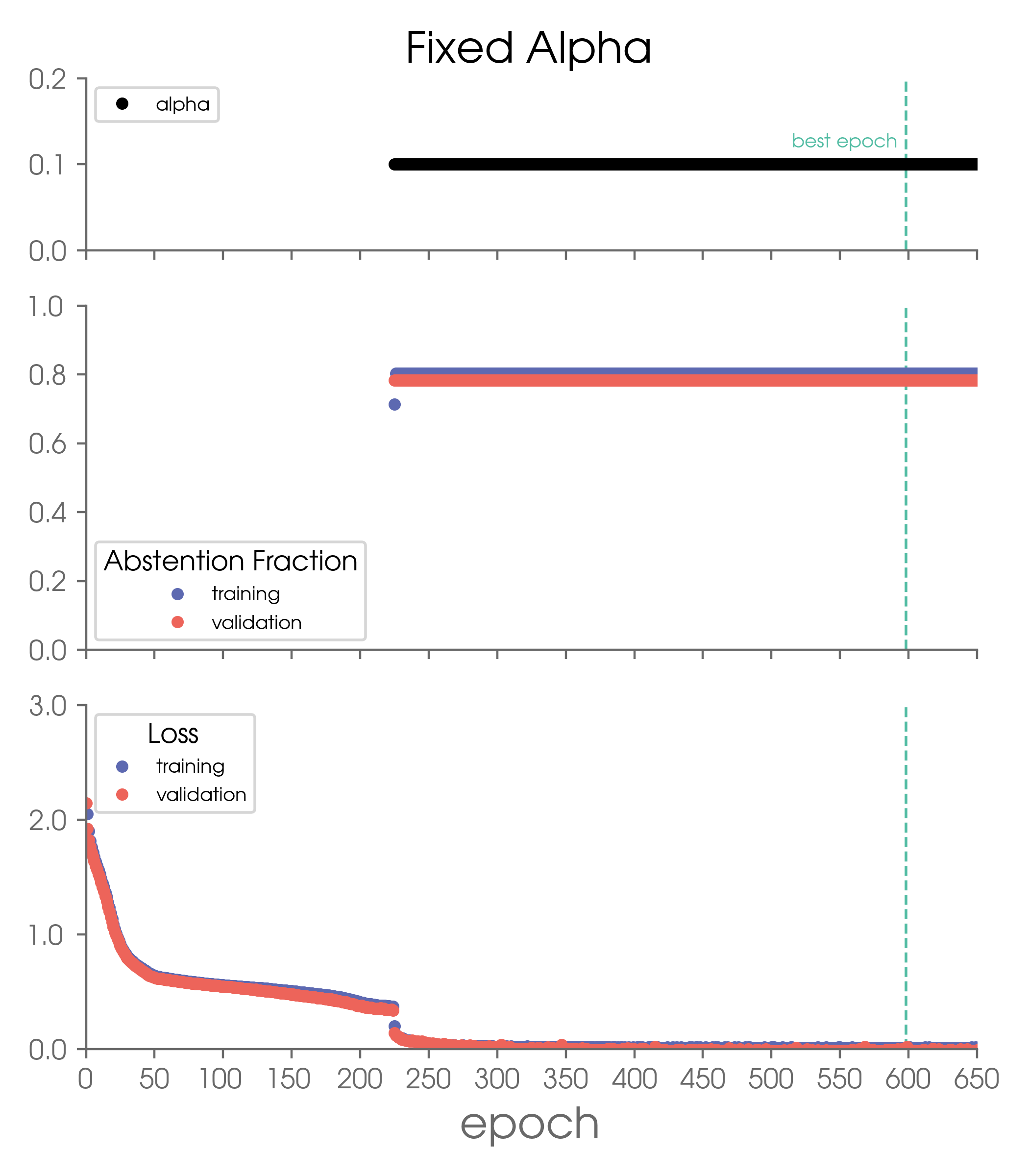}
\end{center}
\caption{\textbf{1D Example w/ Constant $\alpha$.} Example training and validation metrics for a constant $\alpha=0.1$.}
\label{fig_ols_epochs}
\end{figure}

Fig. \ref{fig_ols_epochs} shows $\alpha$ (fixed to 0.1 after spin-up), the abstention fraction, and the loss as a function of epoch during training for one particular model. The loss of both the training and validation data drops steadily during the spin-up stage of 0-225 epochs. At the start of the abstention stage, $\alpha$ is fixed to 0.1, while the abstention fraction is allowed to vary. However, it is clear that the network identifies an optimal abstention fraction by the second epoch of the abstention stage, and network does not vary for the rest of the training. Training is halted by early-stopping and the best weights are taken from the best model at epoch 559.

\begin{figure}
\begin{center}
\noindent\includegraphics[width=200px]{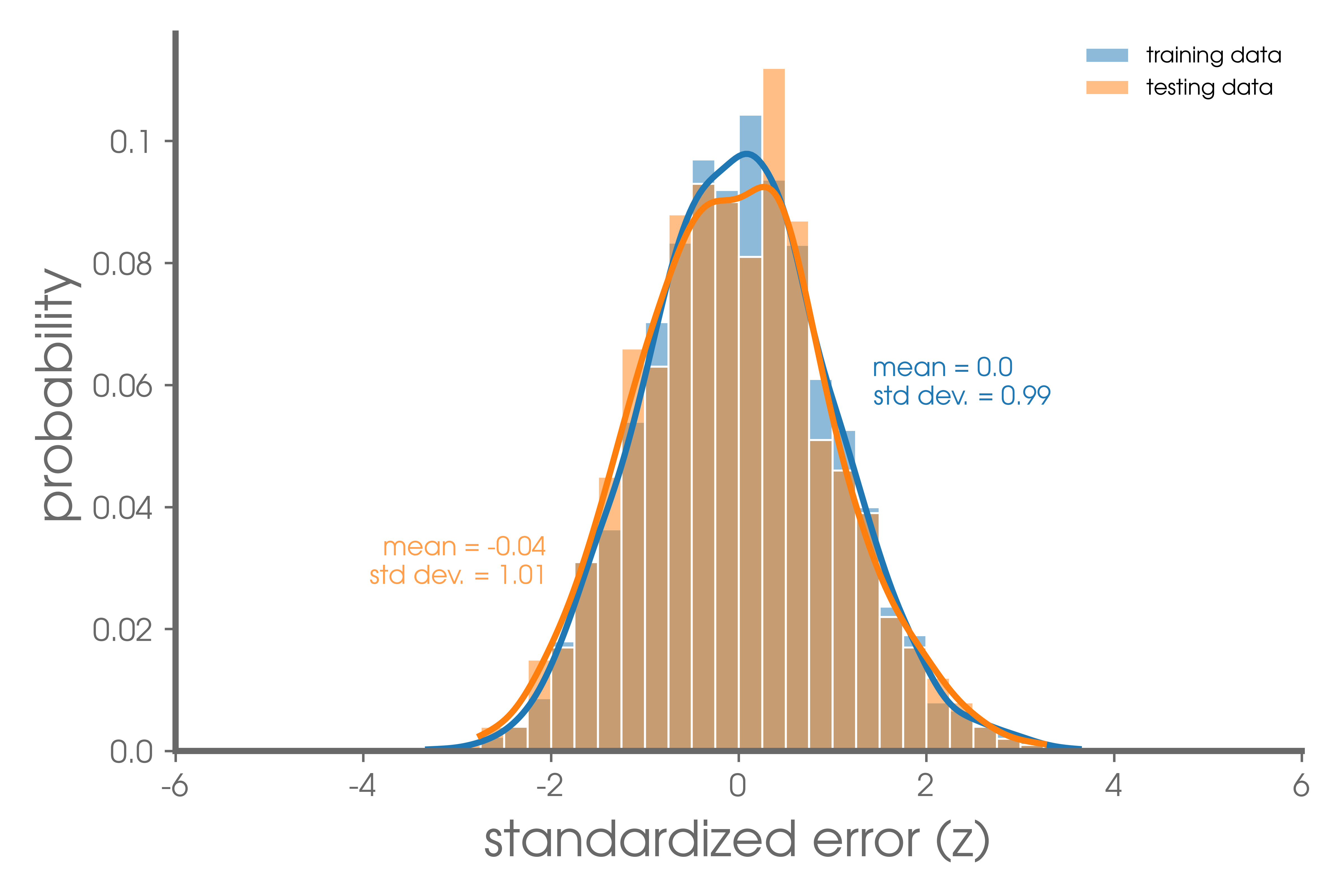}
\end{center}
\caption{\textbf{1D Example w/ Constant $\alpha$.} Histograms of the standardized errors (z-scores) of predictions by the baseline ANN for all samples. Means and standard deviations of these standardized errors are shown in colored text.}
\label{fig_ols_hist}
\end{figure}

Results from the baseline ANN and the CAN are shown in Fig. \ref{fig_ols_summary}b,c,d. As shown in Fig. \ref{fig_ols_summary}b,d, the baseline ANN outperforms the MAE model for all coverage percentages. Unlike the MAE model, the baseline ANN learns which samples are more certain and scales its predicted $\sigma$ accordingly, as shown by the inset histogram in Fig. \ref{fig_ols_summary}b. Fig. \ref{fig_ols_hist} shows the histograms of the standardized errors $z_i$ for the baseline ANN, which are defined as
\begin{linenomath*}
\begin{equation} \label{zj}
z_i = \frac{y_i - \mu_i}{\sigma_i}.
\end{equation}
\end{linenomath*}
The mean and standard deviation of the $z_i$ are approximately $0$ and $1$ for both training and validation. This reveals that the $\sigma$ are more than just unscaled measures of relative confidence.  Rather, we may usefully interpret $\mu_i$ and $\sigma_i$ as the mean and standard deviation of an approximate conditional probability distribution for prediction $j$.

Focusing more closely on the baseline ANN results in Fig. \ref{fig_ols_summary}d (solid purple line), the error decreases with the coverage percent.  This indicates that the more confident predictions are also more correct. As mentioned in the introduction, this is the idea behind \textit{forecasts of opportunity}, and the baseline ANN alone is able to identify the most skillful forecasts without abstention. Even so, the CAN (orange dots) outperforms the baseline ANN slightly: its error is slightly below even the best baseline ANN model and does a slightly better job learning the best fit line (Fig. \ref{fig_ols_summary}c). The CAN obtains its edge over the baseline ANN because it is able to put even more energy into learning the relationships of the confident samples because of the abstention loss design. Furthermore, recall that 20\% of the data falls along the well-defined line in Fig. \ref{fig_ols_summary}a, and the CAN is able to identify the optimal coverage percent as 19\%.

\subsection{Forecasts of Opportunity}
For our first use case with the synthetic climate data, we modify the data to loosely reflect forecasts of opportunity related to teleconnections associated with the El Ni\~no Southern Oscillation (ENSO). Warm ENSO events (El Ni\~no events) have long been known to impact global temperatures and precipitation \cite<e.g.>{McPhaden2006-pi,Yeh2018-tf}.  At times these events have led to skillful forecasts on subseasonal-to-seasonal time scales \cite<e.g.>{Johnson2014-fh}. To mimic this behavior with our synthetic data set, we average the anomalous SSTs in the ENSO region within the equatorial eastern Pacific (dashed white box in the map in Fig. \ref{fig_arch}). When the average value in this box is larger than 0.5 (29\% of the samples), we leave the sample as is. This reflects an opportunity where a strong El Ni\~no may lead to more predictable behavior of the global climate system. Samples where the average value is less than 0.5 represent ``noisy'' samples consequently, we shuffle the $y$ values across these samples so that there is no relationship between the input maps $x$ and their labels $y$. With such a setup, we anticipate that the network can identify strong synthetic El Ni\~no samples (i.e., large values within the ENSO box, Fig. \ref{fig_arch}) as samples with high confidence and low error.

\begin{figure}
\begin{center}
\noindent\includegraphics[width=185px]{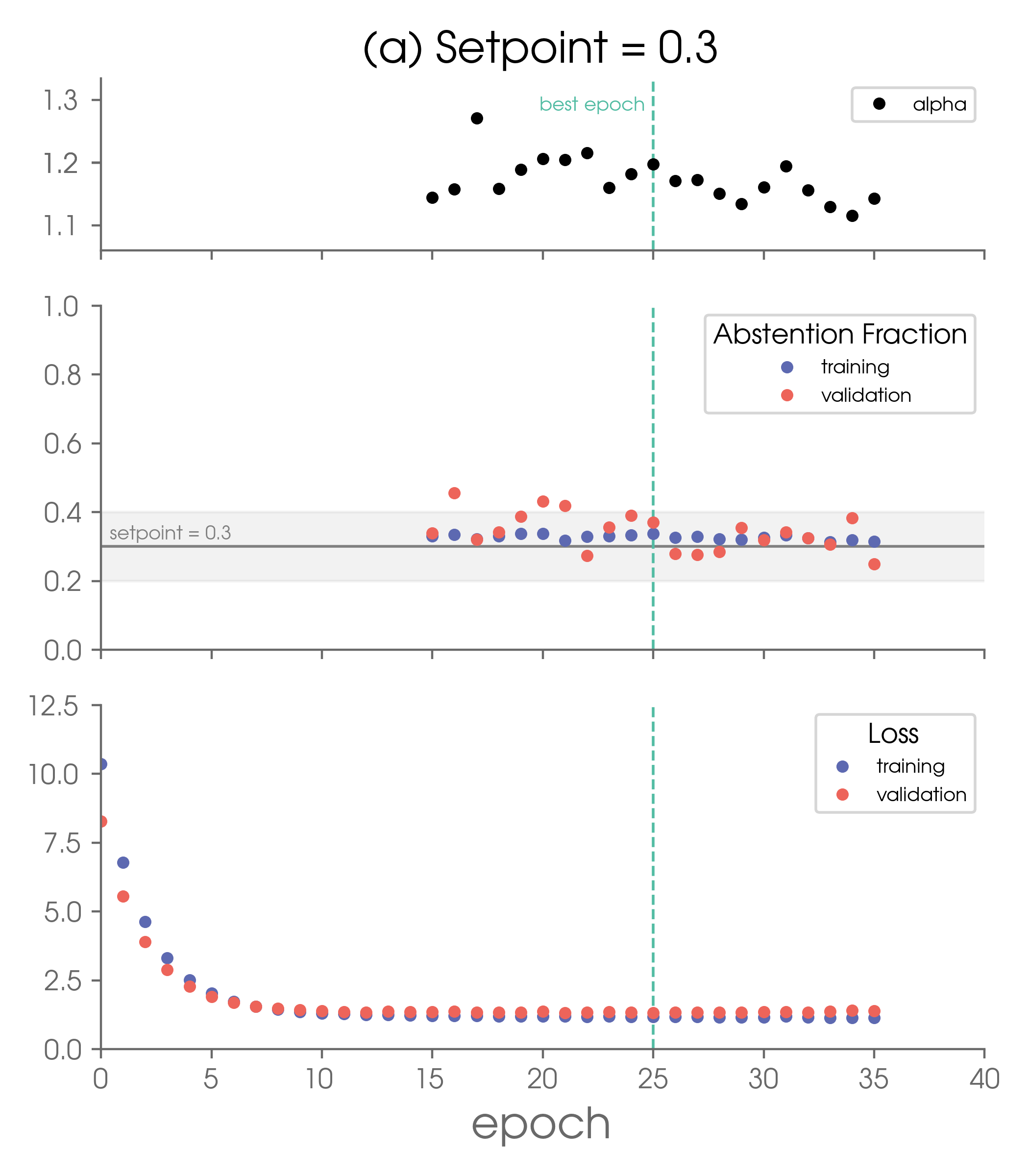}
\noindent\includegraphics[width=185px]{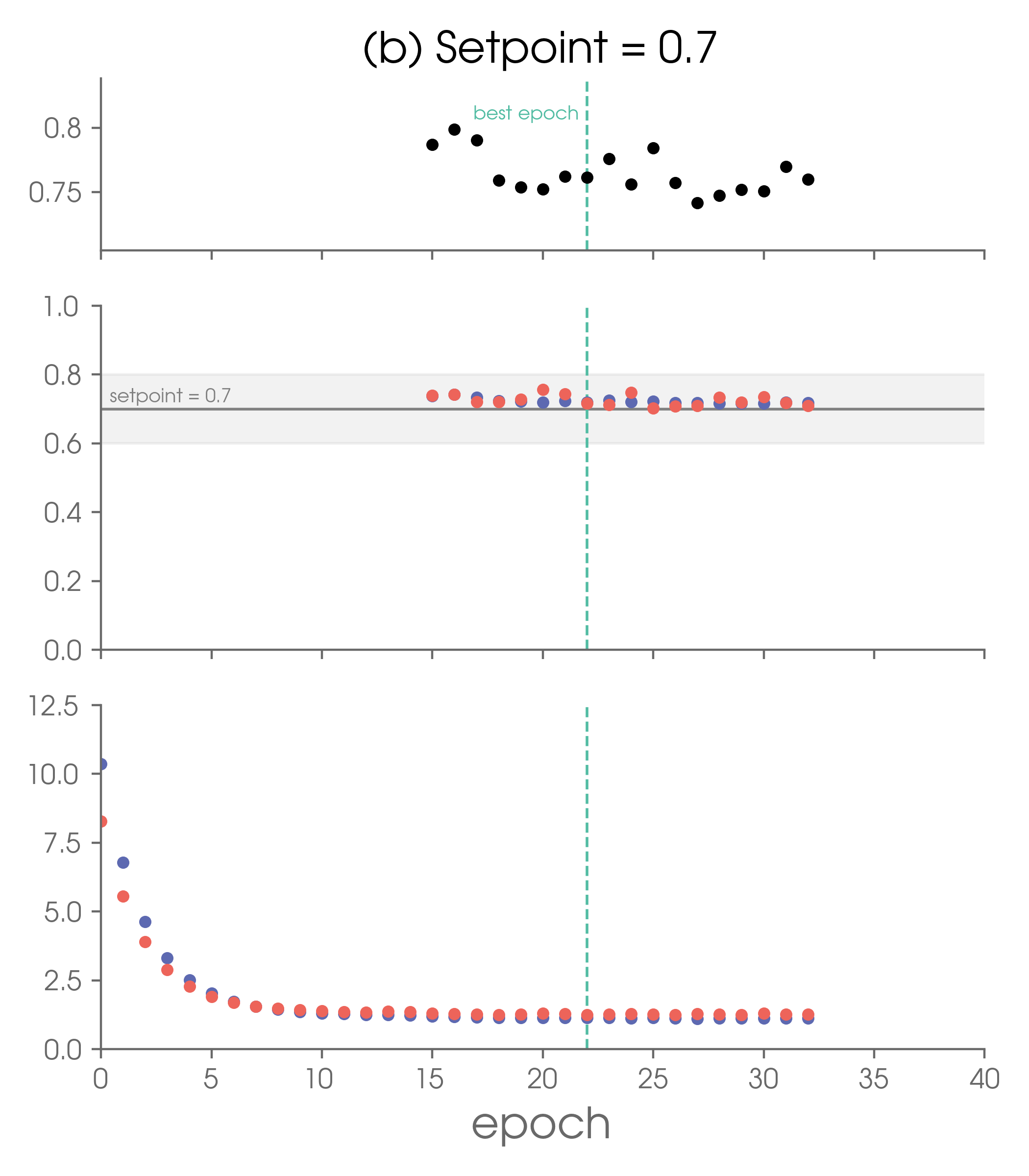}
\end{center}
\caption{\textbf{Forecasts Of Opportunity Experiment w/ PID-controlled $\alpha$.} Example training and validation metrics for abstention setpoints of (a) 0.3 and (b) 0.7.}
\label{fig_epochs}
\end{figure}

We train separate models for abstention setpoints ranging from .1 to .9 in increments of 0.1. Fig. \ref{fig_epochs} shows $\alpha$, the abstention fraction, and the loss as a function of epoch during training for two different abstention setpoints. Following the spin-up period of 15 epochs, the PID-controller adjusts $\alpha$ to maintain an abstention fraction within 0.1 of the abstention setpoint. 

\begin{figure}
\begin{center}
\noindent\includegraphics[width=400px]{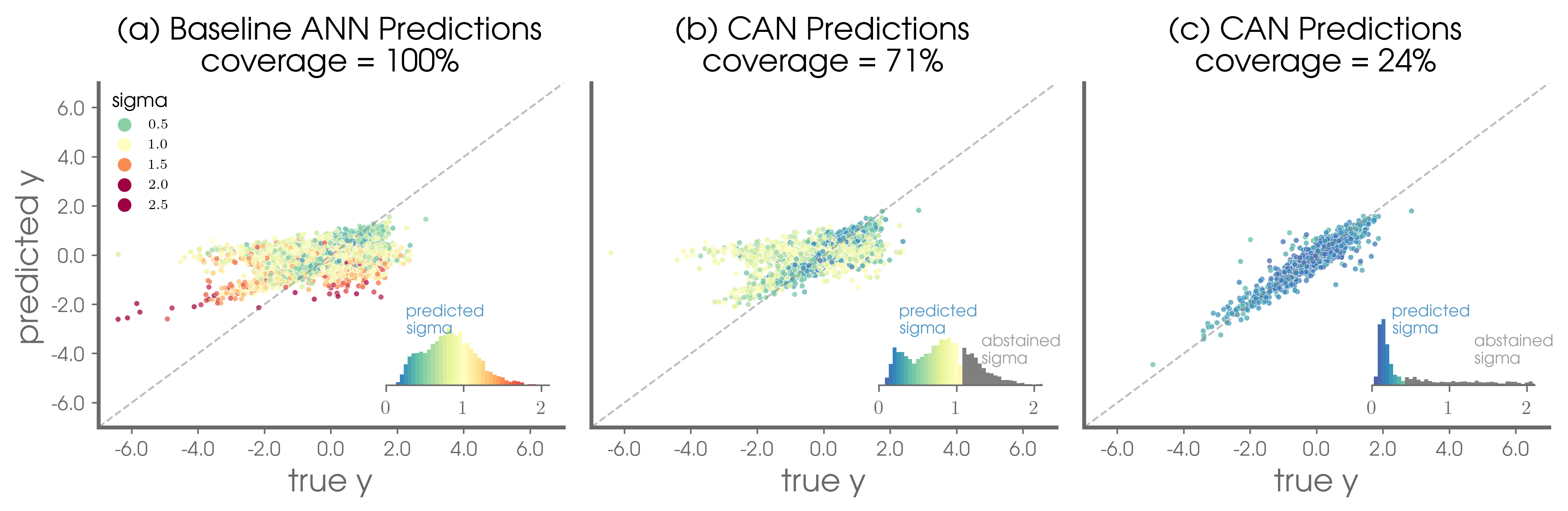}
\noindent\includegraphics[width=400px]{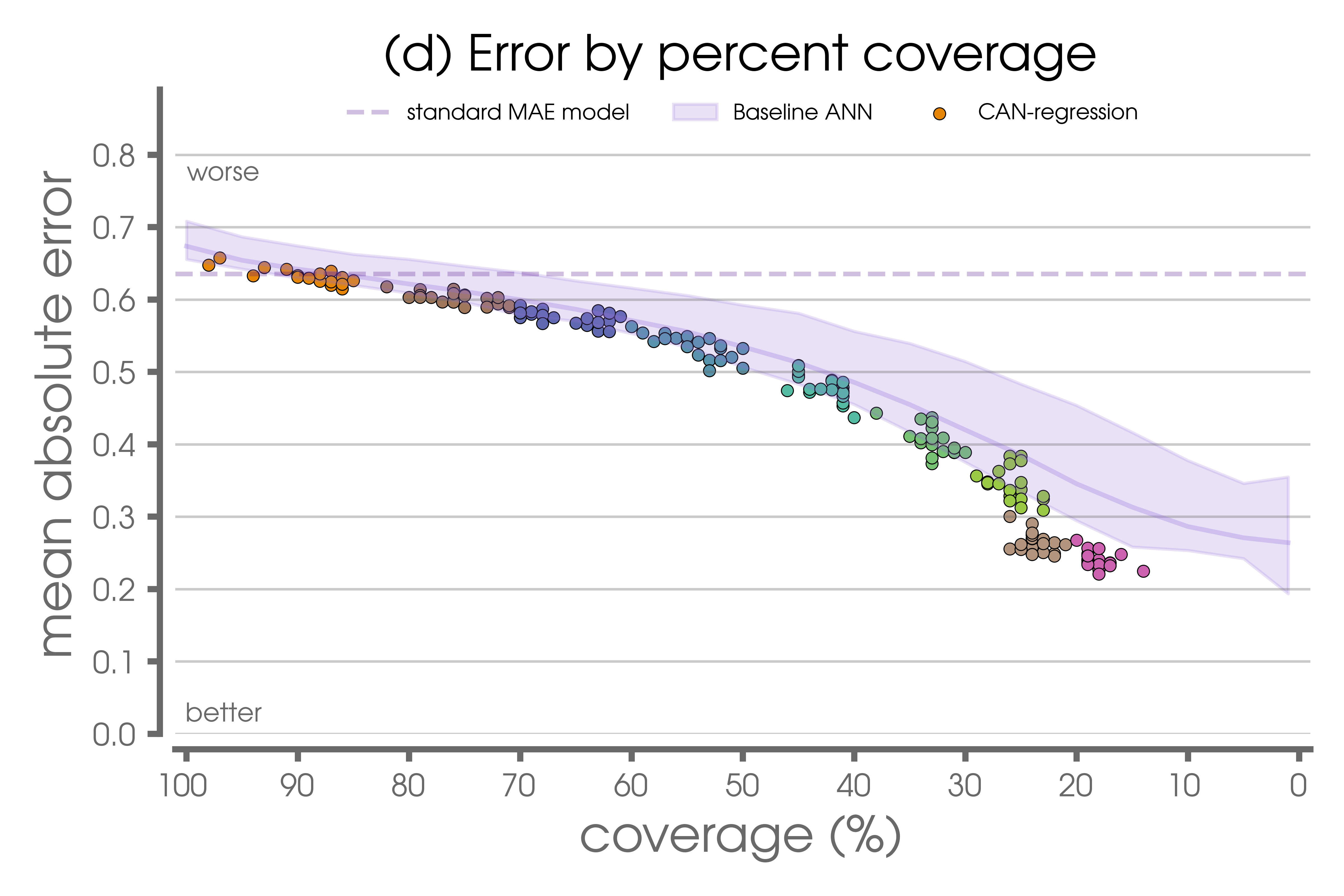}
\end{center}
\caption{\textbf{Forecasts Of Opportunity Experiment w/ PID-controlled $\alpha$.} (a) Predicted $y$ versus the true $y$ for the baseline ANN predictions. The dashed line denotes the one-to-one line --- a perfect prediction. (b,c) are the same (a), but for CAN predictions at two different coverage rates. Scatter plots only show covered predictions (i.e., non-abstained). Colors in (a)-(c) denote the predicted $\sigma$, and insets in (a)-(c) display histograms of the predicted $\sigma$ for both covered and abstained predictions. (d) Mean absolute error versus coverage for different neural network loss functions over a range of initialization seeds and abstention setpoints (shown in colors). Purple shading denotes the full range of errors over 20 baseline ANN models; the solid purple line denotes the median.}
\label{fig_summary}
\end{figure}

Results for the baseline ANN and PID-controlled CAN are shown in Fig. \ref{fig_summary}. As shown in Fig. \ref{fig_summary}d, the baseline ANN error (purple shading) decreases for decreasing coverage. This documents the ability of the baseline ANN to identify the forecasts of opportunity while it assigns higher $\sigma$ values to samples with higher uncertainty (Fig. \ref{fig_summary}a). The colored dots in Fig. \ref{fig_summary}d show results from the PID-controlled CAN for a range of abstention setpoints. Like the baseline ANN, the CAN error decreases with decreasing coverage; however, the best CAN models are always better (lower error) than the best baseline ANN models. This is especially evident for coverage fractions below 30\%, which corresponds to the 29\% of samples that are forecasts of opportunity (i.e., unshuffled). Fig. \ref{fig_summary}b,c display the predictions by the CAN, including histograms of $\sigma$, for two coverage fractions. For lower coverage fractions (higher abstention fractions), the CAN pushes the abstained $\sigma$ values to larger values and likewise improves its confidence on the covered samples by reducing $\sigma$ (compare predicted $\sigma$ histograms inset in Fig. \ref{fig_summary}b,c). That is, the CAN with 24\% coverage learns the forecasts of opportunity samples \textit{better} than the baseline ANN, and better than it does for higher coverage fractions.

\begin{figure}
\begin{center}
\noindent\includegraphics[width=200px]{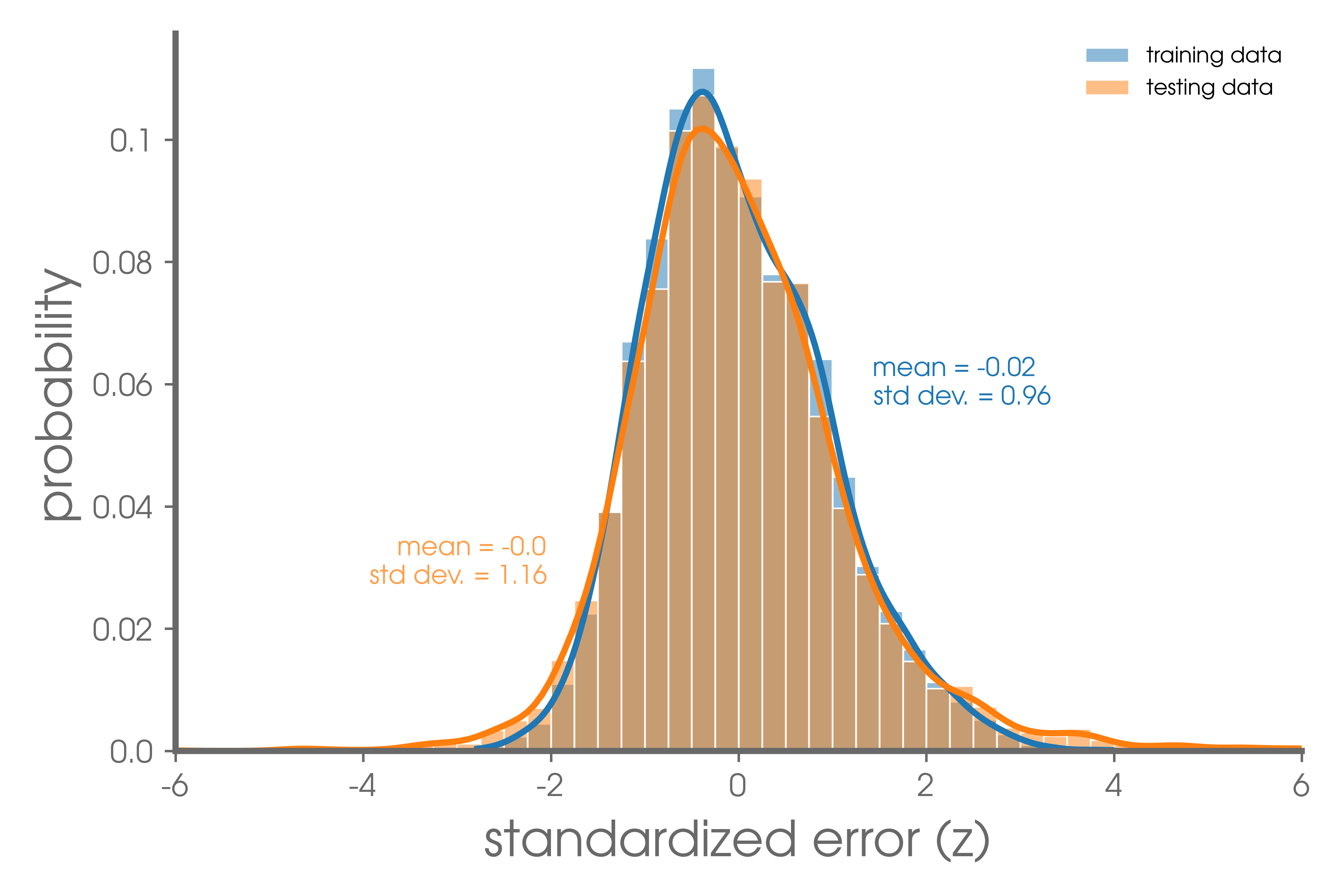}
\end{center}
\caption{\textbf{Forecasts Of Opportunity Experiment w/ PID-controlled $\alpha$.} Histograms of the standardized errors (z-scores) of the predictions by the baseline ANN for all samples. Means and standard deviations of these standardized errors are shown in colored text.}
\label{fig_hist}
\end{figure}

Fig. \ref{fig_hist} shows the histograms of the standardized errors $z_i$ from the baseline ANN (see Eq.~\ref{zj}). As in the 1D example, the mean and standard deviation of the $z_i$ are approximately $0$ and $1$ for both training and testing data (validation data looks similar). This reveals that the $\sigma$ are more than just unscaled measures of relative confidence.  Moreover, we may usefully interpret $\mu_i$ and $\sigma_i$ as the mean and standard deviation of an approximate conditional probability distribution for prediction $j$. One can also create histograms for the CAN of the predicted samples (not shown); however, in this case the histograms are much narrower since the covered (non-abstained) samples tend to be highly confident and exhibit small $\sigma$, as expected.

\begin{figure}
\begin{center}
\noindent\includegraphics[width=266px]{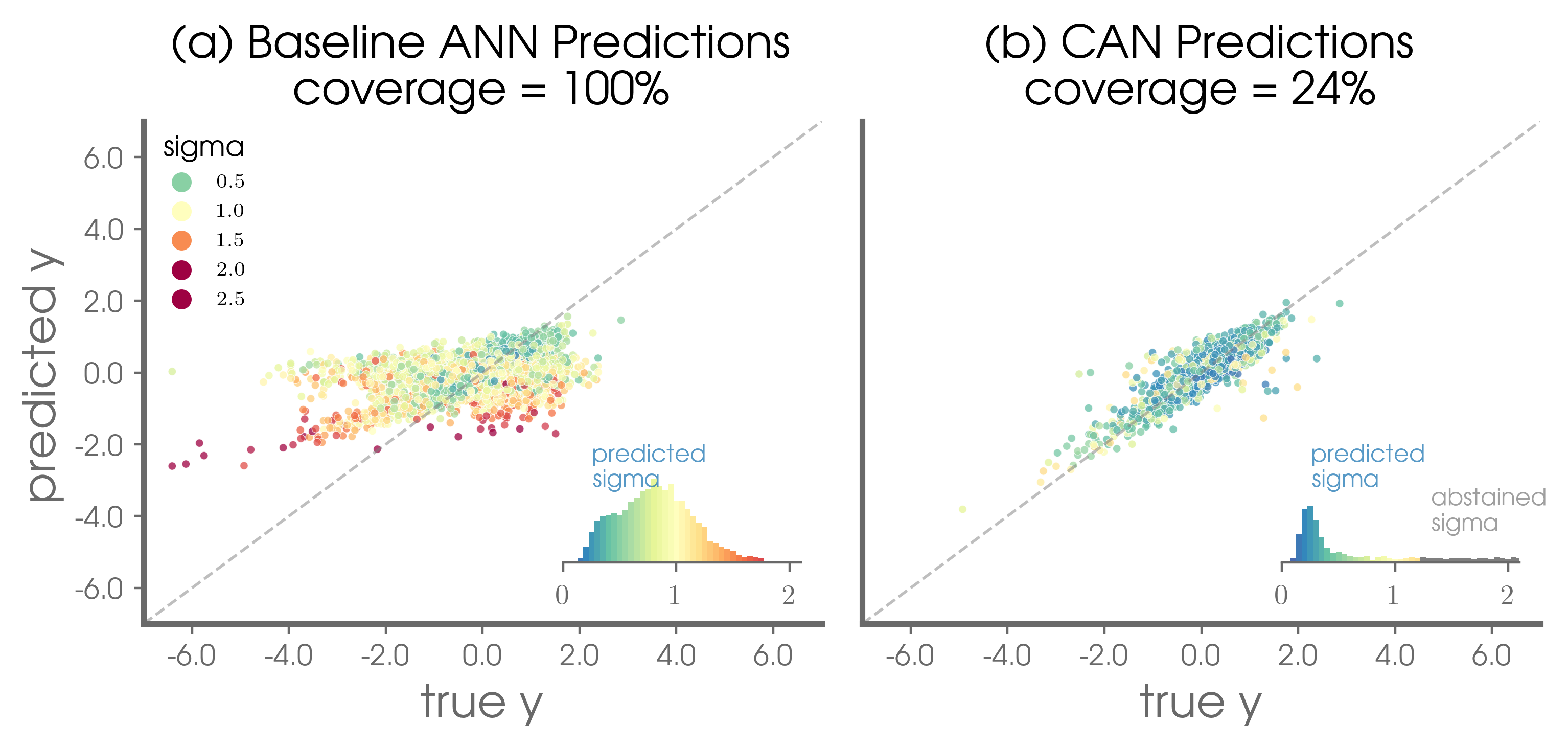}
\noindent\includegraphics[width=400px]{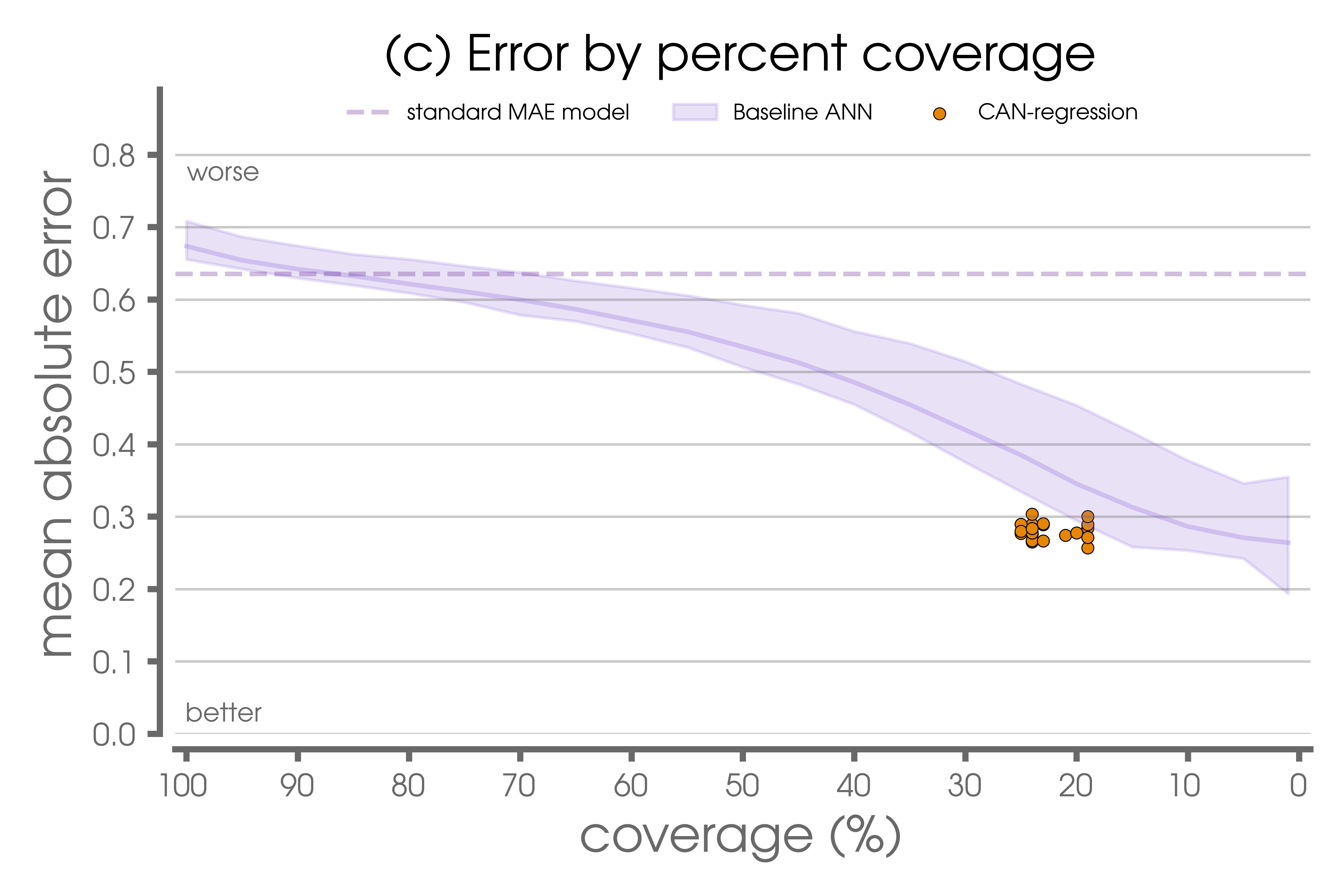}
\end{center}
\caption{\textbf{Forecasts Of Opportunity Experiment w/ Constant $\alpha$.} (a) Predicted $y$ versus the true $y$ for the baseline ANN predictions. The dashed line denotes the one-to-one line --- a perfect prediction. (b) As in (a) but for CAN predictions at a coverage rate of 24\%. Scatter plots only show covered predictions (i.e., non-abstained). Colors in (a,b) denote the predicted $\sigma$, and insets in (a,b) display histograms of the predicted $\sigma$ for both covered and abstained predictions. (d) Mean absolute error versus coverage for different neural network loss functions over a range of initialization seeds for constant $\alpha=0.1$. Purple shading denotes the full range of errors over 20 baseline ANN models; the solid purple line denotes the median.}
\label{fig_const_summary}
\end{figure}

Thus far, we have trained the CAN to identify synthetic El Ni\~no forecasts of opportunity with the PID-controller, which sets the abstention setpoint during training. We can instead use the constant $\alpha$ approach to see if the CAN identifies the correct abstention fraction. We set $\alpha=0.1$; the results are shown in Fig. \ref{fig_const_summary}. The CAN outperforms the baseline ANN with constant $\alpha$, as it did with the PID-controller. In addition, it identifies a coverage of $\sim 24$\%, which is very close to the 29\% forecasts of opportunity samples. 

Interestingly, we find that the PID-controller method tends to slightly outperform the constant $\alpha$ approach (compare the 25\% coverage errors between Fig. \ref{fig_summary}d and \ref{fig_const_summary}c). It is unclear to the authors why this is the case; it could be a function of this synthetic data set. Future work will explore this behavior further.

\subsection{Corrupt Inputs}
\begin{figure}
\begin{center}
\noindent\includegraphics[width=400px]{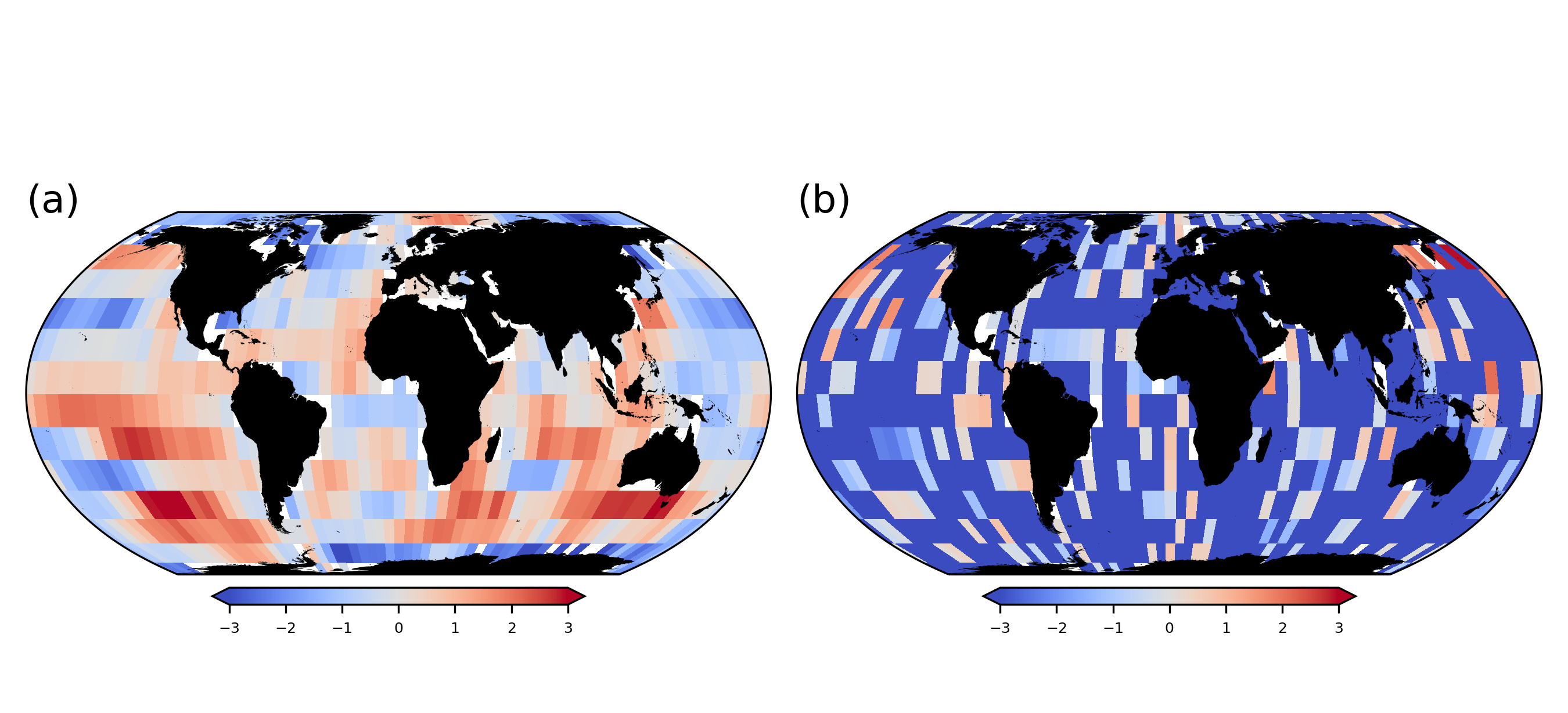}
\end{center}
\caption{\textbf{Corrupt Inputs Experiment.} Examples of (a), an unmodified input map, and (b), a corrupted input map where 66\% of the pixels have been set to $-4.0$.}
\label{fig_corrupt_map}
\end{figure}

\begin{figure}
\begin{center}
\noindent\includegraphics[width=266px]{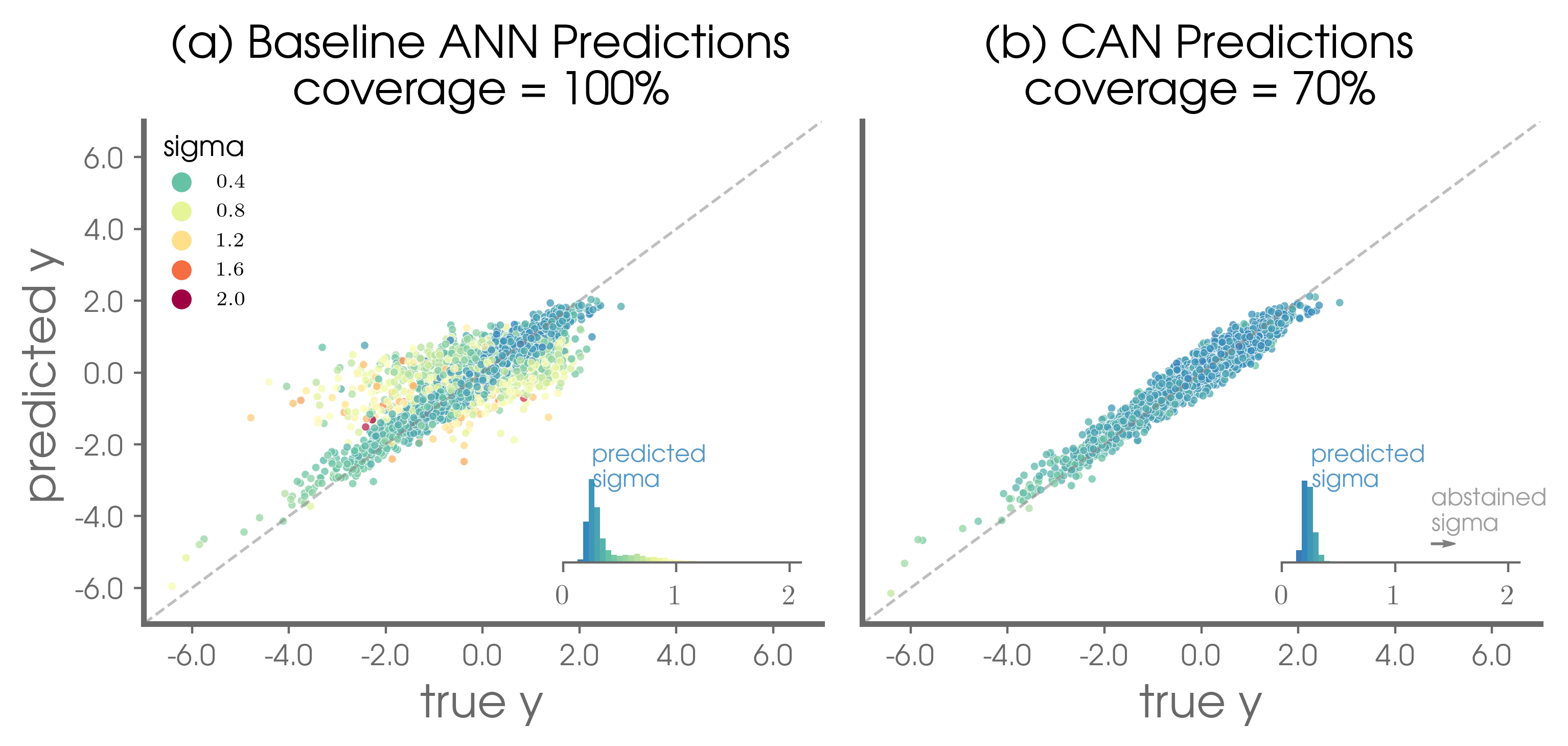}
\noindent\includegraphics[width=400px]{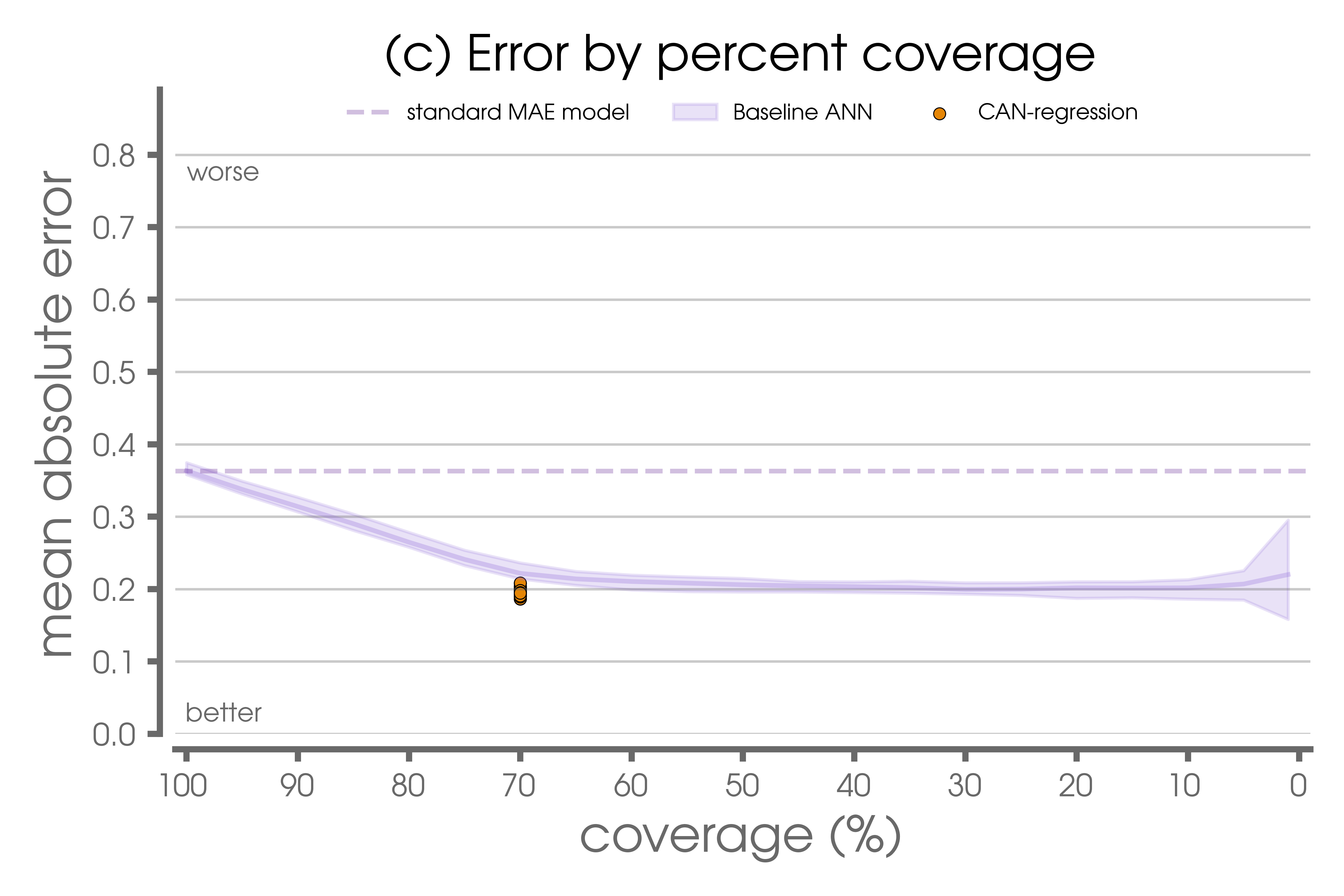}
\end{center}
\caption{\textbf{Corrupt Inputs Experiment w/ Constant $\alpha$.} (a) Predicted $y$ versus the true $y$ for the baseline ANN predictions. The dashed line denotes the one-to-one line --- a perfect prediction. (b) is the same as (a), but for CAN predictions at a coverage rate of 24\%. Scatter plots only show covered predictions (i.e., non-abstained). Colors in (a,b) denote the predicted $\sigma$, and insets in (a,b) display histograms of the predicted $\sigma$ for both covered and abstained predictions. (c) Mean absolute error versus coverage for different neural network loss functions over a range of initialization seeds for constant $\alpha=0.05$. Purple shading denotes the full range of errors over 20 baseline ANN models; the solid purple line denotes the median.}
\label{fig_corrupt_summary}
\end{figure}

For the second synthetic use case, we modify the climate input maps by ``corrupting" some of the grid points by setting them equal to $-4.0$. This exercise is meant to mimic a data set where some of the inputs have bad pixels in some areas. An example of this is shown in Fig. \ref{fig_corrupt_map}. We corrupt 30\% of the samples and leave the remaining 70\% unmodified.  We use the CAN with constant $\alpha=0.05$ to assess whether the network is able to successfully identify the correct abstention fraction of 30\%. Results are shown in Fig. \ref{fig_corrupt_summary}. Once again, the baseline ANN outperforms the standard MAE model for coverages less than 100\%. Furthermore, the CAN outperforms the baseline ANN and correctly identifies 70\% coverage (30\% abstention) as the optimal fraction.  

This use case demonstrates the ability of the CAN to act as a ``data cleaner'' for regression problems \cite{Thulasidasan2019}; the CAN preferentially learns on the uncorrupted samples and abstains on the corrupted ones. Note that if we had corrupted samples in the training set only (not in the testing set), we could remove these corrupted samples prior to training to obtain a model that performs well on the clean data set. This is different than what we have done here. We have trained the network to not only learn the uncorrupted samples, but to also learn to \textit{identify the corrupted samples}. This means that in the future, when new, unseen samples are pushed through the network, the network will be able to handle them accordingly whether they are corrupted or not.

\section{Discussion}
In many ways, abstention loss is yet another approach to combat overfitting, if we think broadly of overfitting as incorrectly learning ``noise'' within the training samples that is not present in the validation or testing samples. Common approaches for dealing with overfitting include dropout \cite{Srivastava2014-bs} and regularization. To explore this a bit further, we reran our forecast of opportunity experiment shown in Fig. \ref{fig_summary} but applied ridge regression with an L$_2$ parameter of 0.1 \cite{Marquardt1975-ac} to the first layer of the network. Ridge regression reduces the magnitude of individual weights, and thus spreads the importance across multiple units \cite<see>[Fig. 3]{Barnes2020-toe}. Results, shown in Supp. Fig. 1, can be directly compared with those in Fig. \ref{fig_summary}. Regularization slightly reduces both the baseline ANN and CAN errors, and allows the baseline ANN to perform more similarly to the CAN. Even so, the CAN outperforms the baseline ANN for the lowest coverage fractions, consistent with the fraction of noisy samples within the synthetic data set. Overall, we see that for this specific use case, regularization can be paired with the abstention loss to produce an even better prediction. 

Results presented here were based on the synthetic climate data of \citeA{Mamalakis2021}, where each sample is independent and the input and output values are largely symmetric about zero. However, real data seldom behave so well. It is likely that real data may require a transformation (e.g. standardization or a power transformation) prior to training if we are to interpret $\mu_i$ and $\sigma_i$ as the mean and standard deviation of an approximate conditional probability distribution for prediction $j$. Furthermore, a potential concern is that we only present use cases based on synthetic climate data. Our aim in this paper is to demonstrate the basic concept and implementation of the abstention loss in a setting where the correct answer is known. This leaves exploration of CAN's utility in specific scientific contexts to future research. With that said, previous work exploring forecasts of opportunity in observations taking a baseline ANN approach \cite<e.g.>[]{Mayer2020,Barnes2020} provides confidence that the abstention loss will be beneficial. 

While we have shown that the abstention loss outperforms the baseline ANN approach, we wish to stress that this baseline approach is itself a simple yet powerful method for incorporating uncertainty into neural network regression problems. This is especially true because the output offers approximate conditional probability distributions for the predictions. Although this baseline approach is a standard in the computer science literature \cite<e.g.,>[Chapters 4 and 5]{Duerr2020}, it much less known in the geoscience community. The authors believe it will be powerful tool as we move forward.

\section{Conclusions}
The ability to say ``I don't know'' is an important skill for any scientist. 

In the context of prediction with deep learning, the identification of uncertain (unpredictable) samples is often approached post-training. In this paper we propose an alternative: a deep learning loss function that can abstain \textit{during training} for regression problems. We first present a baseline regression approach and then introduce a new abstention loss for regression. The abstention loss controlled abstention network (CAN) allows the network to preferentially learn more from confident samples, and ultimately outperform the baseline ANN approach. 

An additional benefit of both the baseline ANN and abstention loss CAN is their simplicity -- they are straightforward to implement in most any network architecture as they only require modification of the output layer and training loss. The abstention loss framework has the potential to aid deep learning algorithms to identify skillful forecasts, as well as corrupt samples, ultimately improving performance on the samples with predictability.

\clearpage

%
%
%
%
%
%
%
%

\acknowledgments
This work was funded, in part, by the NSF AI Institute for Research on Trustworthy AI in Weather, Climate, and Coastal Oceanography (AI2ES) under NSF grant ICER-2019758. Once published, the code and data will be made available to the community via the Mountain Scholar permanent data repository with a permanent DOI and via Zenodo.


%
%

\bibliography{bib_file_abstention.bib}

%
%
%
%
%

\end{document}